\documentclass[runningheads]{llncs}
\usepackage{graphicx}
\usepackage{todonotes}
\usepackage{amsmath}
\usepackage{amsfonts}
\usepackage{subcaption}

\begin{document}
%

\title{Integrating cognitive map learning and active inference for planning in ambiguous environments}

\titlerunning{Active Inference with Cognitive Maps}
%
%
\author{Toon Van de Maele$^1$, Bart Dhoedt$^1$, Tim Verbelen$^{2,}$\thanks{Equal Contribution} and Giovanni Pezzulo$^{3,*}$}
\institute{
    $^1$ IDLab, Department of Information Technology, Ghent University - imec, Belgium\\
    $^2$ VERSES Research Lab, Los Angeles, USA \\
    $^3$ Institute of Cognitive Sciences and Technologies, National Research Council, Italy \\
    \email{toon.vandemaele@ugent.be}
}

\authorrunning{T. Van de Maele et al.}

%
%
\maketitle

\begin{abstract}

Living organisms need to acquire both cognitive maps for learning the structure of the world and planning mechanisms able to deal with the challenges of navigating ambiguous environments. Although significant progress has been made in each of these areas independently, the best way to integrate them is an open research question. In this paper, we propose the integration of a statistical model of cognitive map formation within an active inference agent that supports planning under uncertainty. Specifically, we examine the clone-structured cognitive graph (CSCG) model of cognitive map formation and compare a naive clone graph agent with an active inference-driven clone graph agent, in three spatial navigation scenarios. Our findings demonstrate that while both agents are effective in simple scenarios, the active inference agent is more effective when planning in challenging scenarios, in which sensory observations provide ambiguous information about location. 


\end{abstract}

\section{Introduction} 

Cognitive maps~\cite{okeefe_precis_1979} are mental representations of spatial and conceptual relationships. They are considered essential components for intelligent reasoning and planning, as they are often associated with navigation in humans and rodents~\cite{peer_structuring_2021}. For this reason, a lot of recent developments in both neuroscience and computer science have been building computational models of cognitive maps~\cite{whittington_how_2022}. 

These advances in the field~\cite{whittington_tolman-eichenbaum_2020,george_clone-structured_2021} are very impressive in learning abstract representations and even show that biological patterns such as grid cells~\cite{whittington_tolman-eichenbaum_2020}, or splitter cells~\cite{george_clone-structured_2021} can emerge from learning. However, these works typically do not focus on complex planning tasks and only consider naive or greedy strategies. 

In this paper, we investigate the potential of active inference as a planning mechanism for these cognitive maps. Active inference is a corollary of the free energy principle which states that intelligent agents infer actions that minimize their expected free energy. This is a proxy or bound on expected surprise, yielding a natural trade-off between exploration and goal-driven exploitation~\cite{parr_active_2022,schwartenbeck_computational_2019}. We aim to investigate the impact of active inference as a planning mechanism on the performance of cognitive maps in spatial navigation strategies, especially in terms of disambiguating the ``mental position'' and decision-making efficiency. 

In particular, we look at the clone-structured cognitive graph (CSCG)~\cite{george_clone-structured_2021}: a unifying model for two essential properties of cognitive maps. First, flexible planning behavior, i.e. if observations are not consistent with the expected observation in the plan, the plan can be adapted. Second, the model is able to disambiguate aliased observations depending on the context in which it is encountered, e.g. in spatial alternation tasks at the same location different decisions are made depending on context~\cite{jadhav_awake_2012}. Given the CSCG's inherent mechanism for disambiguating aliased observations, we hypothesize that coupling it with active inference as a planning system will enable the identification of the optimal sequence that accurately represents the agent's location.

To investigate this hypothesized benefit of active inference, we compare both a naive clone graph and an active inference-driven clone graph for navigating toward goals on two separate metrics: the number of steps it takes for an agent to reach the goal and the overall success rate. We design three distinct spatial navigation scenarios, each with a different complexity. First, we consider a slightly ambiguous (open room) environment described by~\cite{george_clone-structured_2021} where we evaluate the structure learning mechanism and planning algorithms for both models. We then increase the level of ambiguity in a maze described in~\cite{friston_sophisticated_2020} where we believe that information-seeking behavior will be crucial for self-localization. Finally, we evaluate the performance in the T-maze, where an agent is punished for making the wrong choice by ending the episode. To summarize, the contributions of this paper are: (i) we show how to use the learned structure of a CSCG as the generative model within the active inference framework, (ii) we show that active inference agents are significantly faster in disambiguating the state in highly ambiguous environments than greedy planning agents, and (iii) we show that active inference agents make more careful decisions by first gathering evidence, yielding higher success rates for finding the reward in the T-maze environment.

\section{Methods}
\label{sect:method}

In this section, we first describe the mechanisms driving standard clone-structured cognitive graphs for structure learning. Then we provide a brief summary of the active inference framework and how the action is driven through Bayesian inference. Finally, we conclude this section by showing how the CSCG can be used as a generative model within the active inference framework.  

\subsection{Clone-Structured Cognitive Graphs}
\label{sect:cscg}

\begin{figure}[t!]
    \centering
    \hfill
    \begin{subfigure}{0.35\textwidth}
        \includegraphics[width=\textwidth]{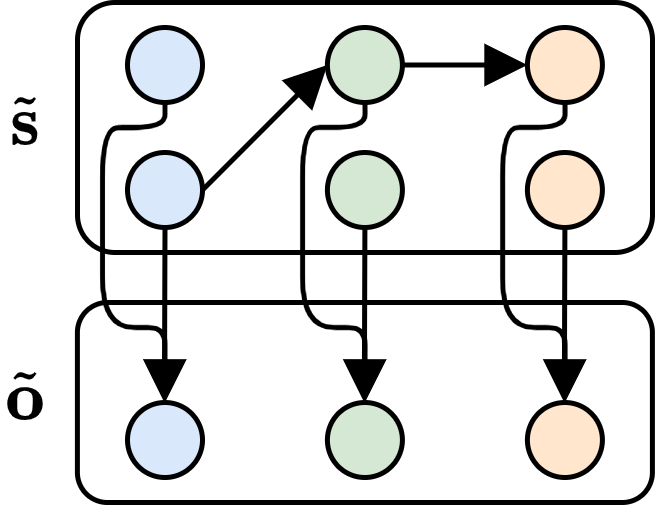}
        \subcaption[]{}
    \end{subfigure}
    \hfill
    \begin{subfigure}{0.5\textwidth}
        \includegraphics[width=\textwidth]{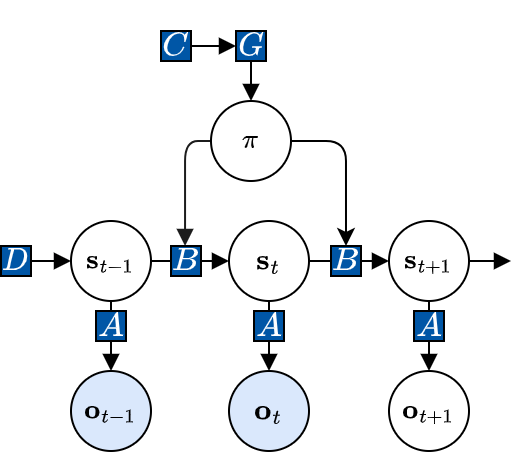}
        \subcaption[]{}
    \end{subfigure}
    \hfill
    \caption{(a) A mapping of a sequence of observations to distinct clone states in the clone-structured cognitive graph. The color indicates clones belonging to a specific observation, i.e. for each colored observation there are two clones states from which it can transition into either clone state belonging to the next observation. (b) The factor graph describing an active inference driven partially observable Markov decision process (POMDP). $\pi$ denotes the policy, which is sampled according to the expected free energy $G$, dependent on the preference matrix $C$. The hidden states of the agent $\mathbf{s}_t$ are initialized using the prior matrix $D$. These states are then transitioned according to the $B$ matrix, conditioned on the selected policy. Finally, the observed outcome variables are generated through the likelihood factor ($A$ matrix). Observed variables are denoted in light blue circles, while unobserved variables are denoted in white circles. The factors describing the generative model are denoted in a dark blue square.} 
    \label{fig:generative_model}
\end{figure}

Clone-structured cognitive graphs (CSCG)~\cite{george_clone-structured_2021} are a computational implementation of a cognitive map that models the joint probability of a sequence of action and observation pairs. They are a variation of the action-augmented hidden Markov model, where the next state and action are conditioned on the current state and action. The crucial difference is that these clone-structured cognitive graphs are able to disambiguate aliased observations based on the context (e.g. the previously visited trajectory), which is a property that is also observed in hippocampal splitter cells. 


In order for a CSCG to be able to disambiguate observations, it needs distinct states for each observation based on its context - in this case, the previous observations and actions. All states corresponding to a single observation are called the clones of this observation, and by design, each state deterministically maps to a single observation. In essence, a CSCG is a hidden Markov model in which multiple different values of the hidden state predict identical observations (i.e. their corresponding columns in the transition matrix are non-identical). A pair of the clone states in a CSCG is therefore a set of two values that a hidden state might take which share identical likelihood contingencies, but differ in their transition probabilities. A depiction of the clone graph, as described in~\cite{george_clone-structured_2021} is shown in Figure~\ref{fig:generative_model}a. 

The CSCGs are optimized by minimizing the variational free energy over a sequence of observation-action pairs using the Baum-Welch algorithm~\cite{emalgo}, an expectation-maximization scheme for hidden Markov models. Through this optimization and random initialization, the model will converge to use distinct clone states for different sequences in the data. This distinction between clones is further improved by optimizing the learned model parameters through a Viterbi decoding step, only keeping the states necessary for the maximum likelihood paths in the learned model. 


\subsection{Clone graph agent}
\label{sect:cga}

We define a clone graph agent that uses a greedy planning approach to select the actions. Planning using the clone-structured cognitive graphs is done by setting a fixed target state (or states), and forward propagating the messages starting from the current state. When one of the target states is assigned a non-zero probability, a path is found and the maximum likelihood states are backward propagated to retrieve the corresponding action sequence, or policy. The probability of each policy is computed as the belief over the current state $Q(\mathbf{s}|\tilde{\mathbf{o}},\tilde{\mathbf{a}})$. Once the agent's belief over state collapses to a single state, the planning mechanism falls back to the one described in~\cite{george_clone-structured_2021}, where the current state is known.

\subsection{Active inference agent}
\label{sect:aif}

Actionable agents, whether biological or artificial, are separated by their environment through sensory inputs (perception) and action. The agent's observations are indirectly observed through its different sensory modalities, while the world state is also only indirectly affected by the agent's actions. This separation between the hidden variables (action, observation, agent state, and world state) is commonly referred to as the Markov blanket. 

The free energy principle proposes that an agent possesses a generative model that describes how outcomes are generated from the world state and how the world state is affected by the agent's actions. The principle states that the agents will minimize their surprise, bounded by the variational free energy by updating the parameters of the generative model (learning) or inferring the hidden state (perception). Active inference agents can infer the action that minimizes the ``expected free energy ($G$)'' (or in other words, the free energy of the future courses of actions)~\cite{parr_active_2022}. 


Active inference assumes that actions are inferred through the minimization of the expected free energy $G$. This means that the posterior over a policy is proportional to the expected free energy $G$, which can be computed for each policy. More specifically, approximate posterior over policy $Q(\pi)$ is computed as the softmax ($\sigma$) over the categorical over all the policies with a value of the respective expected free energy $G$, $\gamma$ is a temperature variable:
$$
    Q(\pi) = \sigma(-\gamma G(\pi)), 
$$
Where the expected free energy G of this model, for a fixed time horizon $T$, is defined as in~\cite{heins_pymdp_2022}:
$$
    G(\pi) = \sum_{\tau=t+1}^T G(\pi,\tau)
$$
$$
    G(\pi,\tau) \geq - \underbrace{\mathbb{E}_{Q(\mathbf{o}_\tau|\pi)}\big[D_{KL}[Q(\mathbf{s}_\tau|\mathbf{o}_\tau,\pi)||Q(\mathbf{s}_\tau|\pi)]\big]}_{\text{Epistemic value}} - \underbrace{\mathbb{E}_{Q(\mathbf{o}_\tau|\pi)}\big[\log P(\mathbf{o}) \big]}_{\text{Pragmatic Value}}
$$
This equation decomposes in two distinct terms: an epistemic value computing the information gain term over the belief over the state, and a pragmatic value (or utility) term with respect to a preferred distribution over the observation $P(\mathbf{o})$. In active inference, the goal of an agent is encoded in this prior belief as a preference. In a CSCG, planning is done by setting a preferred state, whereas in active inference this is typically done by setting the preferred observation. In order to make both approaches comparable, here we always plan by setting preferred states (and assume an identity mapping between the state and observation).


Evaluating the expected free energy $G$ for all the considered policies is exponential w.r.t. the time horizon $T$. This limits the tree depth to low values for which this is practically computable. To mitigate this limitation, we set the preference for each state proportional to the distance toward the goal state (in the cognitive map). While this system simplifies computing the utility to be sufficient for a depth of one, the planning mechanism still requires larger depths for achieving (non-greedy) long-term information-seeking behavior. 


\subsubsection{CSCG as the generative model for active inference} 
\label{sect:cscggenmod}

We consider active inference in the discrete state space formulation~\cite{da_costa_active_2020}, as shown in the factor graph in Figure~\ref{fig:generative_model}b. The generative model is therefore described by a set of four specific matrices: the $A$ matrix defines the likelihood model, or how observations are generated from states: $P(\mathbf{o}|\mathbf{s})$, the $B$ matrix defines the transition model, or how the belief over state changes conditioned on an action $\mathbf{a}_t$: $P(\mathbf{s}_{t+1}|\mathbf{s}_t,\mathbf{a}_t)$. The $C$ matrix describes the preference of the agent $P(\mathbf{s})$, and finally, the $D$ matrix describes the prior belief over the initial state $P(\mathbf{s})$.

First, we learn the world structure using a CSCG through the minimization of the evidence lower bound with respect to the model parameters as described in~\cite{george_clone-structured_2021}. We then map the parameters of the learned hidden Markov model to the four matrices describing the active inference model.  

First, we reduce the model by only considering the states for which the transition probability marginalized over action and next state $\sum_\mathbf{s} \sum_\mathbf{a} p(s_t|\mathbf{s},\mathbf{a})$, assuming a uniform distribution over $\mathbf{s}$ and $\mathbf{a}$, is larger than the threshold of $0.0001$. The $A$ matrix can be directly constructed by setting $P(\mathbf{o}_i|\mathbf{s}_j) = 1$ for all remaining clones $\mathbf{s}_j$ of observation $\mathbf{o}_i$.

To construct the $B$ matrix, the transition matrix from the trained CSCG can be taken directly. A crucial difference between the POMPD in discrete time active inference and the CSCG is that the actions are state-conditioned in the latter. This means that starting in some states, an action can not be taken. In the learned transition matrix, the following condition does not always hold: $\sum_{\mathbf{s}_{t+1}} P(\mathbf{s}_{t+1}|\mathbf{s}_t, \mathbf{a}_t) = 1$. We convert this transition matrix to proper probabilities by adding a novel dispreferred state $\mathbf{s}_d$, for which we set the transition probability to 1 in these illegal cases, and for which this state transitions to itself for each possible action. We then normalize the transition matrix such that probabilities sum to 1. We also add a $P(\mathbf{o}_d|\mathbf{s}_d) = 1$ mapping in the $A$ matrix.

The preference of the agent, or $C$ matrix, is not present in the standard formulation of the CSCG. However, the agent is able to plan toward a goal that is set in state space. We model this by setting a preference over this state, or set of states in case of an observation-space preference or multiple target goals. Additionally, for the newly added state $\mathbf{s}_d$ to which the illegal actions are mapped, we set a very low value (as if it would drive you to a state that is farther away from the goal than the maximum distance) in order to drive the agent to avoid these actions when planning according to its expected free energy. 

The prior distribution over the initial state, matrix $D$, is initialized as a uniform prior over all the states. The agent thus starts with no knowledge about the state it is in and has to gather evidence to change this belief.

\section{Results}
\label{sect:results}

In this work, we compare the behavior of two agents that select their actions using a CSCG: the former (``clone graph'', Section~\ref{sect:cga}) agent plans using a greedy approach, whereas the latter (``active inference'', Section~\ref{sect:aif}) agent uses active inference and expected free energy to plan ahead. We also compare these two agents with a random (``random'') agent baseline. In particular, we look at goal-driven behavior in three distinct environments each requiring a different level of information-seeking behavior. First, we consider an open room as proposed in~\cite{george_clone-structured_2021} in which the agent has to reach a uniquely defined corner, for which the goal is provided as a goal observation. Second, we consider a more ambiguous environment in which the agent has to reach the uniquely defined center of a room, but it first needs to localize itself within the room. Finally, we evaluate the approach on the T-maze, where the agent should first observe a cue, as a wrong decision is ``fatal''. 

In each experiment, we first train the generative models as CSCGs and then convert them to discrete state space matrices for active inference within the PyMDP framework~\cite{heins_pymdp_2022}.

\subsection{Navigating in an open room environment}
\label{sect:openroom}

In this first experiment, we investigate the performance of all agents in a simple environment where we hypothesize that there is no immediate gain in using the active inference framework for information-seeking behavior. As the clone graph agent is still able to integrate observations to improve its belief over its current state, we expect both agents to gather enough evidence to accurately plan toward the goal.

\begin{figure}[t!]
    \centering
    \begin{subfigure}[b]{\textwidth}
        \includegraphics[width=\textwidth]{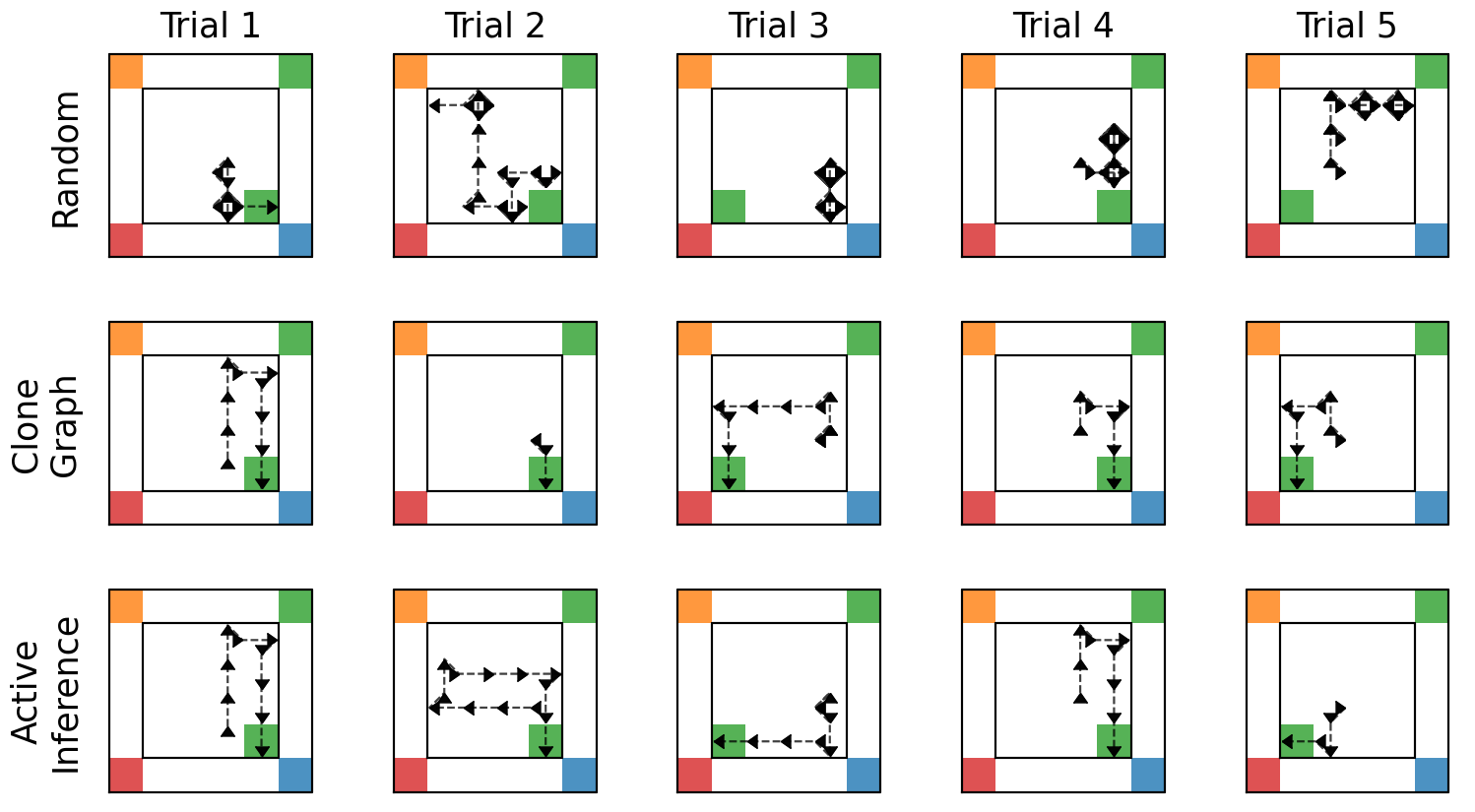}
        \subcaption[]{}
    \end{subfigure}
    \begin{subfigure}[b]{\textwidth}
        \begin{subfigure}[b]{0.12\textwidth}
            \includegraphics[width=\textwidth]{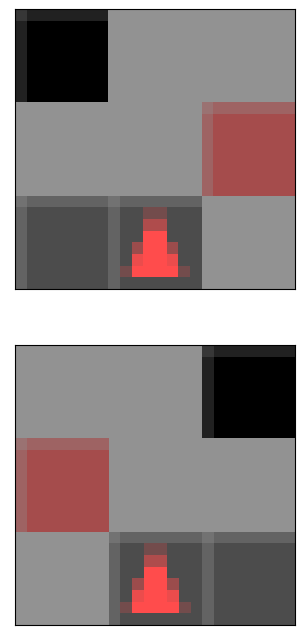}
            \subcaption[]{}
        \end{subfigure}
        \hfill
        \begin{subfigure}[b]{0.42\textwidth}
            \includegraphics[width=\textwidth]{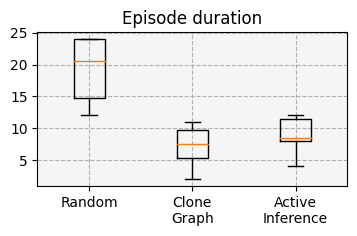}
            \subcaption[]{}
        \end{subfigure}
        \hfill
        \begin{subfigure}[b]{0.42\textwidth}
            \includegraphics[width=\textwidth]{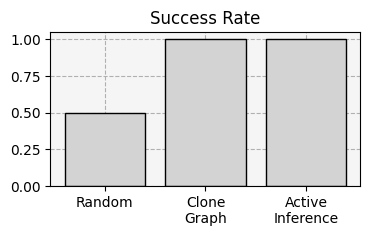}
            \subcaption[]{}
        \end{subfigure}
    \end{subfigure}
    \caption{(a) Qualitative results of navigating the open room maze for the different agents with different random seeds. The agent is tasked with reaching a particular corner in the maze. The trajectory of the agent is marked, and the arrow points the direction in which the agent is looking. (b) The two three-by-three observations defining a goal in a corner of the open room maze. (c) A box plot representing the statistics of the amount of time until the goal is reached (only the success scenarios are considered) over 400 trials. (d) The success rate of the agent in reaching the goal observation (computed over 400 trials).}
    \label{fig:open_room}
\end{figure}

For this maze, we consider an open room environment based on the one described in~\cite{george_clone-structured_2021}. We recreate the environment within the Minigrid~\cite{minigrid} framework. The room is defined by a four-by-four grid in which the agent can freely navigate by selecting actions like ``turn left'', ``turn right'' or ``move forward''. The agent observes a three-by-three patch around its current position, as shown in Figure~\ref{fig:open_room}b. Each corner of the environment is uniquely defined by an observable colored patch, as shown in Figure~\ref{fig:open_room}a and Figure~\ref{fig:open_room}b. Each observed patch is mapped to a unique index as observation. In this environment, this corresponds to 21 observations.

We learn the structure of the room by first training a CSCG, initialized with 20 clones for each observation, as described in Section~\ref{sect:method}. The model parameters were learned using a random-walk sequence consisting of 100k observation-action pairs. We then set the preference of the agent to the two observations reaching the corner, e.g. for the bottom right corner this is the observation of reaching it from the left and from the top. As described in Section~\ref{sect:method}, we select the clone states for which the likelihood of this observation is 1 and set the preference for all these states for both the clone graph and active inference planning schemes.

We run an experiment for all three agents where the agent starts in a random (ambiguous, i.e. looking at the center) pose and has to reach a randomly selected corner as the goal. We run this for 400 separate trials, where each trial was seeded with the same random seed, ensuring that the different agents start with the same starting position and goal. We provide the agents with 25 timesteps to reach the goal and report the success rate and episode length for each of the agents. Qualitatively, in Figure~\ref{fig:open_room}a, we observe that the behavior between the clone graph agent and the active inference agent is very similar; it first picks a corner which is either the goal and the episode ends or an informative landmark, and then the agent moves towards the goal. 

Quantitatively, we observe the duration of the episode and see that the average episode length shown in Figure~\ref{fig:open_room}c is significantly larger for the random agent with respect to both the clone graph agent (2-sample independent t-test, p-value=$7.6\cdot10^{-6}$) and the active inference agent (2-sample independent t-test, p-value=$3.6\cdot 10^{-5}$), illustrating that the model has learned the structure of the world and is not moving randomly. Secondly, we observe that the average episode length of the clone graph agent does not significantly differ from the active inference agent (2-sample independent t-test, p-value=$0.237$), illustrating that for this environment the information-seeking behavior does not benefit performance. This is further evidenced by the success rate shown in Figure~\ref{fig:open_room}d, where the performance of both agents does not significantly differ as they are identical at a 100\% success rate.

From this experiment, we conclude that in an environment where the agent can quickly find an unambiguous landmark such as the corners in the open room, both agents have similar performance.

\subsection{Self-localization in an ambiguous maze} 

In the previous environment, the agent was able to quickly self-localize as random actions would easily disambiguate where in the environment they are. In this experiment, we increase the level of ambiguity and evaluate whether the active inference agent is able to self-localize faster than the clone graph agent. 

\begin{figure}[t!]
    \centering
    \begin{subfigure}[b]{\textwidth}
        \includegraphics[width=\textwidth]{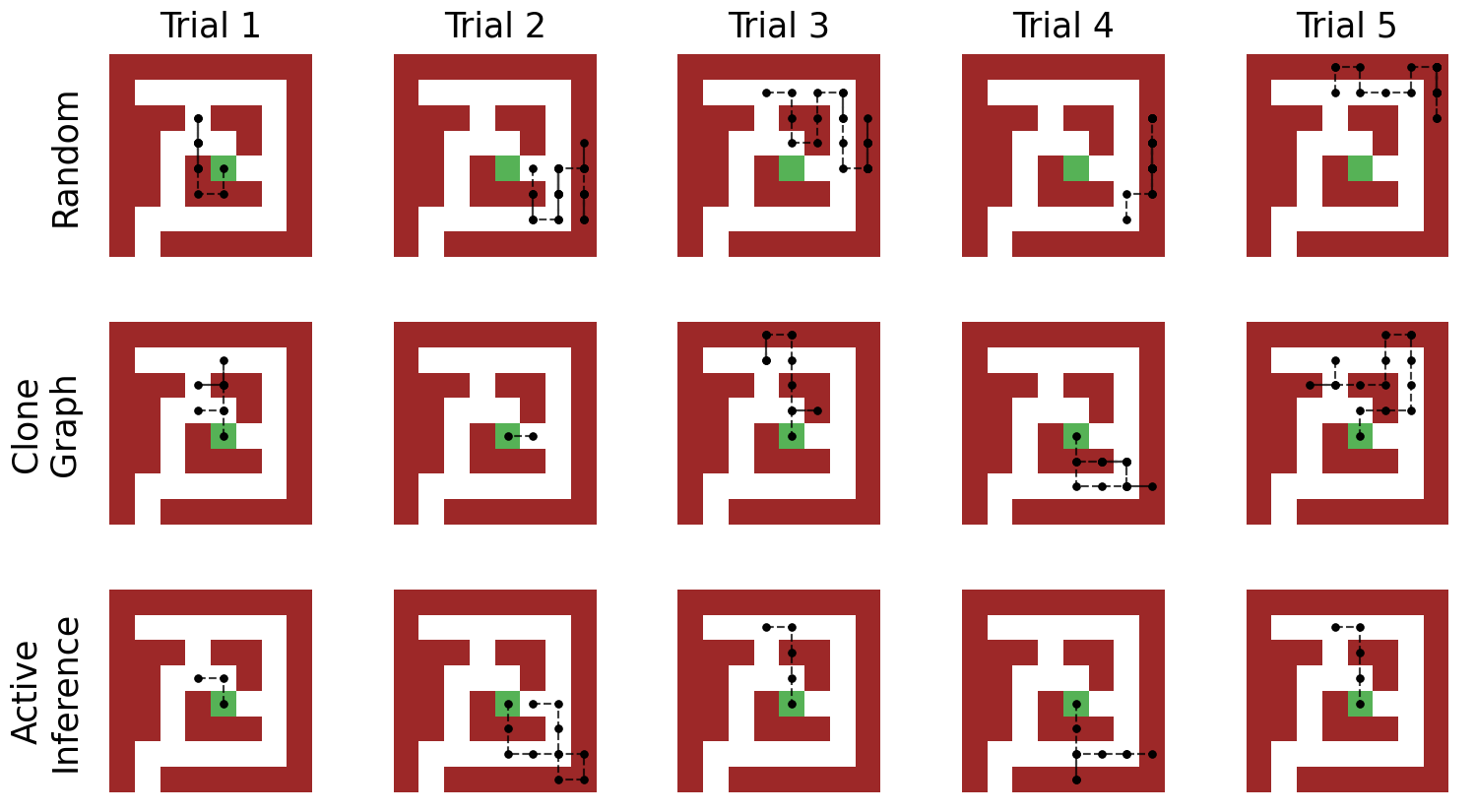}
        \subcaption[]{}
    \end{subfigure}
    \begin{subfigure}[b]{\textwidth}
        \centering
        \begin{subfigure}[b]{0.42\textwidth}
            \includegraphics[width=\textwidth]{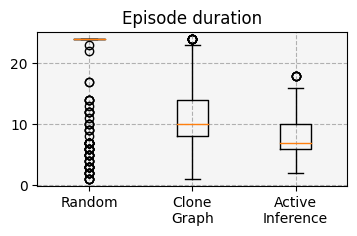}
            \subcaption[]{}
        \end{subfigure}
        \hfill
        \begin{subfigure}[b]{0.42\textwidth}
            \includegraphics[width=\textwidth]{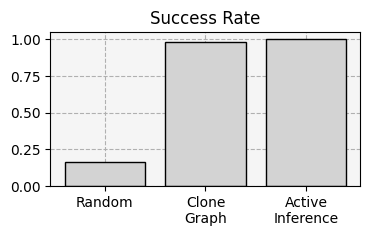}
            \subcaption[]{}
        \end{subfigure}
    \end{subfigure}
    \caption{(a) Qualitative results of navigating the ambiguous maze with the three different agents. The green square marks the goal observation, the trajectory of the agent is marked in black. In this maze, the agent can only observe the current tile, and the color of the tile represents the observation the agent receives. (b) Shows the amount of steps needed for reaching the target, only measured for the success cases. (c) Shows the success rate, computed over 400 trials for the three agents.}
    \label{fig:shockmaze}
\end{figure}

For this experiment, we consider the highly ambiguous maze from Friston et al.~\cite{friston_sophisticated_2020} shown in Figure~\ref{fig:shockmaze}a. In this environment, the agent is only able to observe the one-by-one tile the agent is currently standing on, i.e. if it is a red, white, or green tile. While the red and white tiles are highly ambiguous, there is only a single green tile at the center of the maze. The agent is able to navigate the maze through actions like ``up'', ``down'', ``left'' or ``right'', and is only limited by a wall around the maze. Unique observation tiles are again mapped to categorical indices. 

We construct a CSCG with 40 clones per observation and optimize it over a sequence of 10k steps in the environment until convergence. We then set the preference for this environment as the green tile, in a similar fashion as we did in the experiment in Section~\ref{sect:openroom} for both the clone graph agent and the active inference agent.

In this environment, the agent's goal is always to go to the green tile in the center of the room. However, the agent starts at a random position on a white tile. We again run this experiment for 400 trials for each agent, seeded over trials such that the starting position is the same for each agent. Each episode has a max duration of 25 steps, and we record the episode length and the success rate of the agents. Qualitatively, we can see the trajectories taken by the clone and active inference agents in Figure~\ref{fig:shockmaze}a. We observe that both agents are able to solve the task, seemingly moving randomly in the maze. However, we also observe the random agent navigating in the maze, which typically does not reach the goal. Quantitatively, we again measure that the clone graph agent (2-sample independent t-test, p-value=$1\cdot10^{-99}$) and active inference agent (2-sample independent t-test, p-value=$1\cdot10^{-168}$) significantly differ from the random agent, showing goal-directed behavior. However, we now observe that the clone graph agent with a mean episode duration of $10.92$ steps is significantly slower than the active inference agent with a mean episode duration of $7.92$ steps (2-sample independent t-test, p-value=$3.46\cdot10^{-22}$) even though their success rate is similar with $98.5\%$ for the clone graph agent and $100\%$ for the active inference agent.

From this experiment, we conclude that in highly ambiguous environments, agents using active inference for goal-driven behavior disambiguate their location and reach the goal faster than agents who do not. 

\subsection{Solving the T-Maze}

\begin{figure}[t!]
    \centering
    \begin{subfigure}[b]{\textwidth}
        \includegraphics[width=\textwidth]{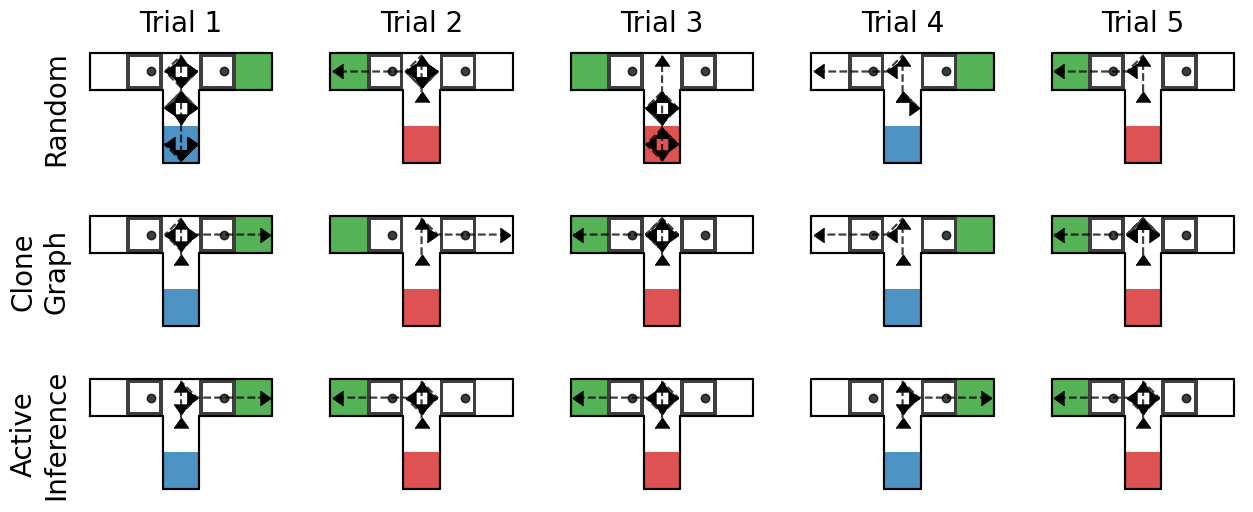}
        \subcaption[]{}
    \end{subfigure}
    \begin{subfigure}[b]{\textwidth}
        \centering
        \begin{subfigure}[b]{0.42\textwidth}
            \includegraphics[width=\textwidth]{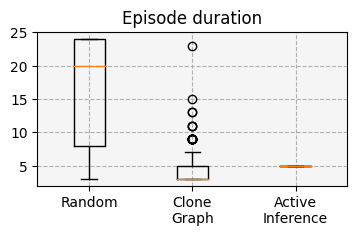}
            \subcaption[]{}
        \end{subfigure}
        \hfill
        \begin{subfigure}[b]{0.42\textwidth}
            \includegraphics[width=\textwidth]{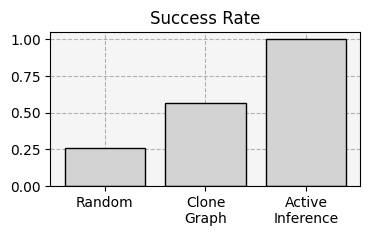}
            \subcaption[]{}
        \end{subfigure}
    \end{subfigure}
    \caption{(a) Qualitative results of navigating the T-maze with the three different agents. The green square marks the goal observation, and the black arrows the trajectory followed by the agent. At the bottom of the T, there is a colored cue, blue marks that the goal is on the right, while red marks that the goal is on the left. (b) Shows the number of steps needed for reaching the target, only measured for the success cases. (c) Shows the success rate, computed over 400 trials for the three agents.}
    \label{fig:tmaze}
\end{figure}

In this final experiment, we consider an environment where making informative decisions is crucial. We compare the performance of the agents in the quintessential active inference environment: the T-maze~\cite{friston_active_2017}. In this environment, the agent must make a choice to go either in the left or the right corridor without being able to observe the location of the reward (we hide it behind a door), and the episode ends when it makes a decision. The agent is, however, able to disambiguate the location of the reward by observing a colored cue behind itself. 

We create the environment again in the Minigrid environment~\cite{minigrid}, and the agent has three-by-three patches as observations and can act by either ``turning left'', ``turning right'' or ``moving forward''. The agent always starts in an upwards-looking position, looking away from the cue. Additionally, when the agent wants to walk through a door, it immediately goes to the tile behind the door, ending the episode either in reward or not. 

We train a CSCG with 5 clones per observation on 500 distinct episodes with a maximum length of 50 steps, however, these episodes are typically shorter as the agent goes through a door. Similar to the open room environment, we map each three-by-three observation patch to a unique index and additionally, we also map the reward to a separate observation. This yields 17 unique observations the agent can observe. We then set the preference to the rewarding observation for both the clone and active inference agents, and depending on context, the agent should be able to infer a different path towards the goal. 

We again conduct 400 random trials, where the seed is again fixed for each trial within an agent, ensuring that for each trial the goal location is the same. When we evaluate the behavior of the agents qualitatively (Figure~\ref{fig:tmaze}a), we observe that the active inference agent always moves forward, turns around and checks for the cue, and then moves towards the correct goal location. In contrast, the clone graph agent randomly picks a direction as it has not accurately inferred in which state it currently is. Interestingly, when the stochasticity of the action sampling forces the agent to turn around and it observes the cue, it chooses the correct action. This explains the 56.75\% success rate, which is slightly higher than the expected 50\% of selecting actions randomly. In this environment, where thoughtless decisions are punished, the active inference agent is significantly more accurate with a success rate of 100\% (2-sample independent z-test for proportions, p-value=$6.25\cdot10^{-50}$). Interestingly, the clone graph agent is significantly faster with an average of 4.5 steps than the active inference agent with an average of 5 steps (2-sample independent t-test, p-value=$2.86\cdot10^{-5}$). This is attributed to the fact that the agent does not take the time to observe the cue and moves towards wherever it believes the goal is. 

From this experiment, we conclude that in information-critical decision-making environments using active inference provides a significant benefit over greedy planning strategies.

\section{Discussion} 

We relate our work to representation learning in complex environments. In the context of learning cognitive maps, work has been done that explicitly separates the underlying spatial structure of the environments with the specific items observed~\cite{whittington_tolman-eichenbaum_2020}. While this model does not entail a generative model, other approaches do consider the hippocampus as a generative model~\cite{stoianov_hippocampal_2022} and show that through generative processes novel plans can be created. Model-based reinforcement learning systems learn similar world models directly from pixels~\cite{hafner_dream_2020} and are able to achieve high performance on RL benchmarks. All these approaches typically treat planning as a trivial problem that can be solved through forward rollouts, or by value optimization using the Bellman equation, however, they do not consider the belief over the state as a parameter.

Within the active inference community, a lot of work has been applied to planning in different types of environments. Casting navigation as inferring the sequence of actions under the generative model using deep neural networks has been done before in~\cite{catal_learning_2020,catal_robot_2021}, where the approximate posterior is implemented through a variational deep neural network. The active inference framework has also been successful in solving various RL benchmarks~\cite{tschantz_scaling_2019,fountas_deep_2020}. These approaches show that inferring action through surprise minimization is powerful in solving a wide range of tasks, although they do not explicitly deal with aliasing in observations.

We believe that the combination of both approaches can yield a promising avenue for building cognitive maps in silico that can be used to solve important real-world tasks such as navigation.


The CSCG has been shown to be a powerful model for flexible planning and disambiguating aliased observation, making it the perfect candidate for integration within the active inference framework. Through this interaction with the inherent uncertainty-resolving behavior of active inference, we have observed significant improvements in terms of success rate or episode lengths depending on the specific environment. 

Another open issue that we plan to resolve in the future is the fact that the CSCG is currently learned in an offline fashion. Therefore our current approach is not benefitting from the curiosity- or novelty-based scheme of active inference~\cite{kaplan_planning_2018,schwartenbeck_computational_2019}, which we hypothesize to improve the training efficiency with respect to the number of required samples.

\section{Conclusion} 

We first propose a mechanism for using the clone-structured cognitive graph within the active inference framework. This allows us to use the naturally context-dependent disambiguating of aliased observations in the generative model within the active inference framework that naturally will seek the sequence best aligned with this purpose. 
Through evaluation in three distinct environments, we have highlighted the advantages of active inference compared to more simplistic and greedy planning methods. We show that in naturally unambiguous environments, the active inference and clone agents perform similarly in both success rate and time to reach the goal. Additionally, we have observed that the active inference agent exhibits a significantly higher success rate in environments requiring informed decision-making. Finally, we show that in environments where an agent has to make an informed decision, the active inference agent has a significantly higher success rate. These results corroborate the benefits of using an active inference approach. 

\subsubsection*{Acknowledgments}
This research received funding from the Flemish Government (AI Research Program). This research was supported by a grant for a research stay abroad by the Flanders Research Foundation (FWO).

\bibliographystyle{ieeetr}
\bibliography{references,extra_references}

\begin{thebibliography}{10}

\bibitem{okeefe_precis_1979}
J.~O'Keefe and L.~Nadel, ``Précis of {O}'{Keefe} \& {Nadel}'s \textit{{The}
  hippocampus as a cognitive map},'' {\em Behavioral and Brain Sciences},
  vol.~2, pp.~487--494, Dec. 1979.

\bibitem{peer_structuring_2021}
M.~Peer, I.~K. Brunec, N.~S. Newcombe, and R.~A. Epstein, ``Structuring
  {Knowledge} with {Cognitive} {Maps} and {Cognitive} {Graphs},'' {\em Trends
  in Cognitive Sciences}, vol.~25, pp.~37--54, Jan. 2021.

\bibitem{whittington_how_2022}
J.~C.~R. Whittington, D.~McCaffary, J.~J.~W. Bakermans, and T.~E.~J. Behrens,
  ``How to build a cognitive map,'' {\em Nature Neuroscience}, vol.~25,
  pp.~1257--1272, Oct. 2022.

\bibitem{whittington_tolman-eichenbaum_2020}
J.~C. Whittington, T.~H. Muller, S.~Mark, G.~Chen, C.~Barry, N.~Burgess, and
  T.~E. Behrens, ``The {Tolman}-{Eichenbaum} {Machine}: {Unifying} {Space} and
  {Relational} {Memory} through {Generalization} in the {Hippocampal}
  {Formation},'' {\em Cell}, vol.~183, pp.~1249--1263.e23, Nov. 2020.

\bibitem{george_clone-structured_2021}
D.~George, R.~V. Rikhye, N.~Gothoskar, J.~S. Guntupalli, A.~Dedieu, and
  M.~Lázaro-Gredilla, ``Clone-structured graph representations enable flexible
  learning and vicarious evaluation of cognitive maps,'' {\em Nature
  Communications}, vol.~12, p.~2392, Apr. 2021.

\bibitem{parr_active_2022}
T.~Parr, G.~Pezzulo, and K.~J. Friston, {\em Active {Inference}: {The} {Free}
  {Energy} {Principle} in {Mind}, {Brain}, and {Behavior}}.
\newblock The MIT Press, 2022.

\bibitem{schwartenbeck_computational_2019}
P.~Schwartenbeck, J.~Passecker, T.~U. Hauser, T.~H. FitzGerald, M.~Kronbichler,
  and K.~J. Friston, ``Computational mechanisms of curiosity and goal-directed
  exploration,'' {\em eLife}, vol.~8, p.~e41703, May 2019.

\bibitem{jadhav_awake_2012}
S.~P. Jadhav, C.~Kemere, P.~W. German, and L.~M. Frank, ``Awake {Hippocampal}
  {Sharp}-{Wave} {Ripples} {Support} {Spatial} {Memory},'' {\em Science},
  vol.~336, pp.~1454--1458, June 2012.

\bibitem{friston_sophisticated_2020}
K.~Friston, L.~Da~Costa, D.~Hafner, C.~Hesp, and T.~Parr, ``Sophisticated
  {Inference},'' 2020.
\newblock Publisher: arXiv Version Number: 1.

\bibitem{emalgo}
C.~F.~J. Wu, ``On the convergence properties of the em algorithm,'' {\em The
  Annals of Statistics}, vol.~11, no.~1, pp.~95--103, 1983.

\bibitem{heins_pymdp_2022}
C.~Heins, B.~Millidge, D.~Demekas, B.~Klein, K.~Friston, I.~Couzin, and
  A.~Tschantz, ``pymdp: {A} {Python} library for active inference in discrete
  state spaces,'' 2022.
\newblock Publisher: arXiv Version Number: 2.

\bibitem{da_costa_active_2020}
L.~Da~Costa, T.~Parr, N.~Sajid, S.~Veselic, V.~Neacsu, and K.~Friston, ``Active
  inference on discrete state-spaces: {A} synthesis,'' {\em Journal of
  Mathematical Psychology}, vol.~99, p.~102447, Dec. 2020.

\bibitem{minigrid}
M.~Chevalier-Boisvert, L.~Willems, and S.~Pal, ``Minimalistic gridworld
  environment for gymnasium,'' 2018.

\bibitem{friston_active_2017}
K.~Friston, T.~FitzGerald, F.~Rigoli, P.~Schwartenbeck, and G.~Pezzulo,
  ``Active {Inference}: {A} {Process} {Theory},'' {\em Neural Computation},
  vol.~29, pp.~1--49, Jan. 2017.

\bibitem{stoianov_hippocampal_2022}
I.~Stoianov, D.~Maisto, and G.~Pezzulo, ``The hippocampal formation as a
  hierarchical generative model supporting generative replay and continual
  learning,'' {\em Progress in Neurobiology}, vol.~217, p.~102329, Oct. 2022.

\bibitem{hafner_dream_2020}
D.~Hafner, T.~Lillicrap, J.~Ba, and M.~Norouzi, ``Dream to {Control}:
  {Learning} {Behaviors} by {Latent} {Imagination},'' Mar. 2020.
\newblock arXiv:1912.01603 [cs].

\bibitem{catal_learning_2020}
O.~Çatal, S.~Wauthier, C.~De~Boom, T.~Verbelen, and B.~Dhoedt, ``Learning
  {Generative} {State} {Space} {Models} for {Active} {Inference},'' {\em
  Frontiers in Computational Neuroscience}, vol.~14, p.~574372, Nov. 2020.

\bibitem{catal_robot_2021}
O.~Çatal, T.~Verbelen, T.~Van De~Maele, B.~Dhoedt, and A.~Safron, ``Robot
  navigation as hierarchical active inference,'' {\em Neural Networks},
  vol.~142, pp.~192--204, Oct. 2021.

\bibitem{tschantz_scaling_2019}
A.~Tschantz, M.~Baltieri, A.~K. Seth, and C.~L. Buckley, ``Scaling active
  inference,'' 2019.
\newblock Publisher: arXiv Version Number: 1.

\bibitem{fountas_deep_2020}
Z.~Fountas, N.~Sajid, P.~A.~M. Mediano, and K.~Friston, ``Deep active inference
  agents using {Monte}-{Carlo} methods,'' Oct. 2020.
\newblock arXiv:2006.04176 [cs, q-bio, stat].

\bibitem{kaplan_planning_2018}
R.~Kaplan and K.~J. Friston, ``Planning and navigation as active inference,''
  {\em Biological Cybernetics}, vol.~112, pp.~323--343, Aug. 2018.

\end{thebibliography}

\end{document}